\documentclass{article}

\usepackage{microtype}
\usepackage{graphicx}
\usepackage{subfigure}
\usepackage{booktabs} %

\usepackage{hyperref}

\usepackage[accepted]{icml2023}

\usepackage{amsmath}
\usepackage{amssymb}
\usepackage{mathtools}
\usepackage{amsthm}
\newcommand\numberthis{\addtocounter{equation}{1}\tag{\theequation}}

\usepackage{dsfont}
\DeclareMathOperator*{\E}{\mathop{\mathbb{E}}}

\let\P\relax

\DeclareMathOperator{\P}{\mathbb{P}}
\DeclareMathOperator{\defeq}{\dot{=}}

\DeclareMathOperator{\ind}{\mathds{1}}

\DeclareMathOperator*{\argmax}{argmax}

\newcommand{\pluseq}{\mathrel{+}=}

\usepackage[capitalize,noabbrev]{cleveref}

\theoremstyle{plain}

\theoremstyle{definition}

\theoremstyle{remark}

\usepackage[textsize=tiny]{todonotes}

\usepackage{multirow}

\icmltitlerunning{Option Iteration}

\begin{document}

\twocolumn[
\icmltitle{Iterative Option Discovery for Planning, by Planning}

\icmlsetsymbol{equal}{*}

\begin{icmlauthorlist}
\icmlauthor{Kenny Young}{UofA} 
\icmlauthor{Richard S. Sutton}{UofA}
\end{icmlauthorlist}

\icmlaffiliation{UofA}{University of Alberta and the Alberta Machine Intelligence Institute}
\icmlcorrespondingauthor{\\Kenny Young}{kjyoung@ualberta.ca}

\icmlkeywords{Machine Learning, ICML}

\vskip 0.3in
]

\printAffiliationsAndNotice{} %

\begin{abstract}
Discovering useful temporal abstractions, in the form of options, is widely thought to be key to applying reinforcement learning and planning to increasingly complex domains. Building on the empirical success of the Expert Iteration approach to policy learning used in AlphaZero, we propose Option Iteration, an analogous approach to option discovery. Rather than learning a single strong policy that is trained to match the search results everywhere, Option Iteration learns a set of option policies trained such that for each state encountered, at least one policy in the set matches the search results for some horizon into the future. Intuitively, this may be significantly easier as it allows the algorithm to hedge its bets compared to learning a single globally strong policy, which may have complex dependencies on the details of the current state. Having learned such a set of locally strong policies, we can use them to guide the search algorithm resulting in a virtuous cycle where better options lead to better search results which allows for training of better options. We demonstrate experimentally that planning using options learned with Option Iteration leads to a significant benefit in challenging planning environments compared to an analogous planning algorithm operating in the space of primitive actions and learning a single rollout policy with Expert Iteration.
\end{abstract}

\section{Introduction}\label{introduction}
Consider the planning problem faced by a human approaching a four-way intersection while driving in an unfamiliar city. Figuring out precisely the correct action to take requires complex reasoning to incorporate knowledge of local geography, as well as the desired destination. However, despite this complexity, there is a relatively small set of short-term temporally extended behaviours which have the potential to be useful. Namely, we may wish to turn left, turn right, or go straight. On the other hand, we can almost always rule out the enormous space of behaviours which would lead us to run off the road. This example highlights the motivation for Option Iteration (OptIt). Namely, the optimal policy may be a very complicated, and thus difficult to learn, function of the current state. Nonetheless, there may be a small set of potentially useful short-term behaviours which are comparatively easy to learn and broadly applicable. 
Once learned, such a set of short-term behaviours, could be used to guide the search toward a smaller set of plausible trajectories. 

Motivated by the above example, OptIt aims to discover a set of options such that in every state visited, at least one option in the set is good for a fixed number of steps into the future. OptIt learns options with an approach similar to Expert Iteration (\citet{anthony2017thinking}; ExIt), and the closely related approach used by AlphaZero~\cite{silver2017mastering}. ExIt is itself essentially a version of approximate policy iteration, a fundamental dynamic programming algorithm. Given some search algorithm which takes a prior policy and uses some computation to generate an improved policy, ExIt iteratively updates the prior policy to better match the output of the search procedure. In doing so, ExIt amortizes the result of the computationally expensive search into a relatively inexpensive neural network. This results in a virtuous cycle in which each search results in an improved policy, which is amortized into a neural network and used to improve future searches. 

Like ExIt, OptIt operates in a decision-time planning setting where a planner has access to a generative model of the environment which can be used to plan in each encountered state before committing to an action. In this work, we assume the ground truth generative model is available, but OptIt could also be applied with a learned model as in, for example, MuZero~\citep{schrittwieser2020mastering}. OptIt applies an ExIt-like procedure to learn a set of policies such that in each encountered state, at least one policy in the set agrees with the result of the search procedure not only in that single state but for a sequence of future states. Intuitively, this is likely easier since it allows the algorithm to hedge its bets. Without learning the nuances of a previously unseen state $s$ well enough to decide which option is best, an algorithm may still be able to discover a set such that at least one option included in the set is likely to be good in $s$. The learned set of options can then be used to improve search by allowing actions to be evaluated under a variety of different plausible behaviours rather than a single learned policy.

\section{Background}
We consider the reinforcement learning problem faced by a goal-directed agent interacting with an initially unknown environment~\citep{sutton2020reinforcement}. We formalize this interaction as a Markov decision process (MDP). An MDP $\mathcal{M}$ is defined by a tuple $\mathcal{M}=(\mathcal{S},\mathcal{A},p,p_0,r,\gamma)$. The agent begins in some state $S_0\in\mathcal{S}$ drawn from the start-state distribution $p_0$. At each time $t\geq0$ an agent observes a state $S_t\in\mathcal{S}$ and based on this information selects an action $A_t\in\mathcal{A}$. Based on $A_t$ and $S_t$, the environment then transitions to a new state $S_{t+1}\in\mathcal{S}$ according to the transition distribution $\P(S_{t+1}=s^\prime|S_t=s,A_t=a)=p(s^\prime|s,a)$, and provides a reward according to the scalar reward function $R_{t+1}=r(S_t,A_t,S_{t+1})$. The discount factor $\gamma\in[0,1]$ defines the time horizon of the problem in a manner that will be explained shortly.

The agent's behaviour is specified by a policy $\pi(a|s)$, which is a distribution over $a\in\mathcal{A}$ for each $s\in\mathcal{S}$. The agent's goal is, roughly, to obtain as much reward as possible in the long run. We will focus on the episodic setting where we assume that, with probability one, for every possible policy the agent environment interaction eventually reaches a terminal state $\bot$ from which no more reward is possible. The termination time $T$ is the random time at which $\bot$ is reached. The discounted sum of rewards obtained from some time $t$ until $T$ is called the return at time $t$, $G_t=\sum\limits_{k=t}^T\gamma^{k-t}R_{k+1}$. The agent's goal is to learn, from a sequence of episodes ending in the terminal state, a policy $\pi$ which maximizes the total expected return $G_0$.

The action-value function of a policy is defined as $q_\pi(s,a)=\E[G_t|S_t=s,A_t=a]$, the expected return if action a is selected in state s and policy $\pi$ is followed from that point forward. The optimal action-value function $q^\star(s,a)=\max\limits_\pi q_\pi(s,a)$ is defined as the maximum action value overall policies $\pi$. Conversely, an optimal policy $\pi^*$ is one such that $q^\star(s,a)= q_{\pi^*}(s,a)$, though note that $\pi^*$ need not be unique.

We consider the case where the agent has access to a generative model of the environment. This means that for any state $s$ and action $a$, the agent can query the model and observe a sampled next state $s^\prime$ from $p(s^\prime|s,a)$ as well the reward $r(s,a)$. The agent can use this model during interaction with the environment to plan for better action selection. While generally cheaper and safer than environment interaction, these queries carry a computational cost and thus we are still concerned with using them efficiently.

Our work builds on the options framework of~\citet{sutton1999between}. In the original formulation, an option $n$ consists of a 3-tuple $(\mathcal{I}_n,\pi_n,\beta_n)$. 
The option's policy $\pi_n(a|s)$ gives the probability of the agent selecting action $a$ in state $s$ while following option $n$. The initiation set $\mathcal{I}_n$ is the set of states where an option can be executed, as is common in the literature we will simply assume all options can be executed everywhere. The termination function, $\beta_n(s)$ gives the probability of option $n$ terminating after entering state $s$. In this work, we will assume options terminate a fixed number of steps after their initiation,\footnote{Technically this makes our options non-Markov as one cannot determine the termination probability from the state alone without knowing the steps since initiation.} but we discuss how this may be extended. Given a set of options, an agent can use them in a manner that is largely interchangeable with actions enabling planning and learning algorithms to run in a temporally abstract space. However, in this work, rather than using options as a drop-in replacement for primitive actions, we will plan in the joint space of primitive actions and subsequent options, allowing each action to be evaluated under a variety of possible future behaviour policies.

\section{Options as a Way to Represent the Joint Distribution Over Future Optimal Actions}\label{joint_action_distribution}
While temporal abstraction is widely regarded as key to scaling reinforcement learning algorithms to increasingly complex problems, empirical results are mixed. 
Likewise, it is not always clear how or why learning a set of options should lead to improvement over simply learning a single strong policy, especially in a single-task setting. For example, \citet{bacon2017option} found that without additional regularization, learned options tend to collapse down to single-step primitive actions. For this reason, before we suggest new techniques for option discovery, it is useful to articulate how specifically we believe learning a set of options could be helpful in a sufficiently complex single-task setting. Here, we expand on the intuition presented in Section~\ref{introduction} to articulate a view that options can allow us to capture information about the joint distribution over optimal actions in a way that learning a single strong policy does not. Our reasoning is reminiscent of the case for learning joint predictions over labels presented by~\citet{osband2021epistemic}, as well as related to posterior sampling for reinforcement learning~\citep{strens2000bayesian,osband2013more}.

Towards formalizing the intuition from Section~\ref{introduction}, define the entropy-regularized optimal policy
\begin{equation}\label{entropy_regularized_optimal_pi}
    \pi_\beta^*(a|s)=\frac{\exp( q^*(s,a)/\beta)}{\sum_{a^\prime}\exp(q^*(s,a^\prime)/\beta))},
\end{equation}
where $q^*(s, a)$ is the unknown optimal action-value function of the true MDP and the entropy-regularization factor $\beta$ is a hyperparameter. Imagine we have observed a dataset $D$ of samples from $\pi_\beta^*$ in various states. Given this data, along with some prior over regularized optimal policies, we can in principle use Bayes theorem to determine a posterior over regularized optimal policies $\P(\pi_\beta^*|\mathcal{D})$.\footnote{Note that while there may be many unregularized optimal policies sharing the same action-values, using a regularized version as in Equation~\ref{entropy_regularized_optimal_pi} makes the solution unique for a given MDP.} Now, imagine we arrive in a new state and would like to provide a planner with an approximation to the distribution of regularized optimal policies to guide its search for a good action in this state. Say we have the choice between specifying this approximation as a single policy, or as a mixture distribution over $N$ different policies. The single policy assigns some probability $\pi(a|s)$, and thus given a sequence of $K$ states $\textbf{s}=(s_0,s_1,....s_{K-1})$ it will assign a joint probability to action sequence $\textbf{a}=(a_0,a_1,....a_{K-1})$ as follows
\begin{equation}\label{single_policy}
    \hat{\P}(\textbf{a}|\textbf{s})=\prod_{k=0}^{K-1}\pi(a_k|s_k).
\end{equation}
On the other hand, given an approximate posterior consisting of a mixture of $N$ policies $\pi_n(a|s)$ for $n\in\{0,...,N-1\}$ weighted by some $\rho(n)$, we get the following joint probability over actions for $\textbf{s}$
\begin{equation}\label{mixture_policy}
    \hat{\P}(\textbf{a}|\textbf{s})=\sum_{n=0}^{N-1}\rho(n)\prod_{k=0}^{K-1}\pi_n(a_k|s_k).
\end{equation}\
It's clear that Equation~\ref{mixture_policy} includes Equation~\ref{single_policy} as a special case, just set $\rho(n)=\ind(n=0)$ and $\pi_0=\pi$. Furthermore, as Equation~\ref{mixture_policy} allows nontrivial correlation between regularized optimal-policy actions in different states, it can in general be a significantly better representation of the true posterior predictive distribution
\begin{equation*}
    \P(\textbf{a}|\textbf{s},\mathcal{D})=\int_{\pi_\beta^*}\P(\pi_\beta^*|\mathcal{D})\prod_{k=0}^{K-1}\pi_\beta^*(a_k|s_k)d\pi_\beta^*.
\end{equation*}
To give a simple example, consider a tabular environment, called Compass, in which an agent starts each episode in a random location on a 2-dimensional, square, grid of cells with $K$ cells in total. In each cell, the agent has the choice to move up, down, left or right. Termination occurs upon reaching one of the 4 edges of the grid, with a positive reward on one edge and a negative reward on the other three. The positive reward is a priori equally likely to lie on any edge. Thus, the optimal policy will always go in the same direction for every state in a particular grid instance. In addition, assume the state includes an integer $m\in\{0,...,M-1\}$ which indicates the identity of the grid. For a given $m$, the positive reward will always be on the same side, but it is a priori unknown which indices correspond to which side. The states of this environment can be represented by $s=(k,m)$ where $k$ indexes all the cells within the grid and $m$ is the index of the grid.

Now assume, based on some combination of prior knowledge and data, the posterior over optimal policies\footnote{We consider unregularized optimal policies in this case.} captures the fact that for a fixed, unobserved, grid the correct behaviour is either to always go left, right, up or down with equal probability. Now, consider approximating the posterior predictive distribution for the set of states $\textbf{s}={\{(k,\tilde{m})\}}_k$ for all $k$ and a single previously unobserved $\tilde{m}$. In particular, consider the minimal cross-entropy approximation using either a single policy as in Equation~\ref{single_policy} or a mixture of policies as in Equation~\ref{mixture_policy}. With a single policy, we aim to minimize
\begin{align*}
    &\E\limits_{\textbf{A}\sim\P(\textbf{A}|\textbf{s},\mathcal{D})}\left[-\log\left(\prod_{k=0}^{K-1}\pi(A_k|(k,\tilde{m}))\right)\right]=\\
    &\sum_{k=0}^{K-1}\E_{A_k\sim\P(A_k|(k,\tilde{m}),\mathcal{D})}\left[-\log(\pi(A_k|(k,\tilde{m})))\right].
\end{align*}
Where $\textbf{A}=(A_0,...,A_{K-1})$ is a random sequence of actions drawn from the true posterior predictive distribution. Since the marginal likelihood $\P(A_k|(k,\tilde{m}),D)$ is $0.25$ for each action, the best we can do is to set $\pi(A_k|(k,\tilde{m}))=0.25$ uniformly. This yields a total cross-entropy of $K\log(4)$. 
On the other hand, using a mixture of four policies we can exactly represent the predictive posterior over ${\{(k,\tilde{m})\}}_k$. In particular, we can set $\rho(0)=\rho(1)=\rho(2)=\rho(3)=0.25$, and $\pi_0(a|(k,\tilde{m})))=\ind(a=left)$, $\pi_1(a|(k,\tilde{m})))=\ind(a=right)$, $\pi_2(a|(k,\tilde{m})))=\ind(a=up)$, $\pi_3(a|(k,\tilde{m})))=\ind(a=down)$ for all $k$. Since this exactly matches the true predictive posterior it gives the best possible cross-entropy of $\log(4)$, $K$ times lower than what is achievable from fitting a single policy, and equal to the true entropy. In turn, this corresponds to a $4^d$ times higher probability of the action sequence that takes the agent directly to the edge with a positive reward, where $d\leq\sqrt{K}$ is the distance from the agents current location to the edge with positive reward. Roughly speaking, without specifying the precise algorithmic details of the planner, a planner using this mixture policy to generate potential action sequences will need to consider exponentially fewer trajectories on average to find the optimal sequence compared to one using the best-fit single policy.

Compass also illustrates the potential benefit of searching for a set of options such that at least one in the set is good only for some short horizon into the future in each state. The same four option policies suffice to represent the posterior predictive distribution over all $s=(k,\tilde{m})$ for any fixed $\tilde{m}$. On the other hand, representing the predictive posterior jointly for all possible $k$ and $m$ would require $4^M$ policies, as we'd have to represent every joint configuration of optimal action for each index $m$. Furthermore, assuming we are using the set of options for planning locally over a particular time horizon, we need not accurately represent the predictive posterior globally, but only locally for the states likely to be encountered during planning.

We postulate that the kind of local structure that exists in Compass, with strong correlation among the optimal actions for temporally contiguous states, is likely to be present in many problems of interest. In particular, due to spatial locality, temporally contiguous states will tend to involve interaction with a highly overlapping set of subsystems in the world and thus any information we gain about these subsystems will be likely to provide information about the optimal behaviour in many such states. Note, however, such local correlation need not hold a priori. One can easily construct posterior distributions such that the optimal action in one state is uninformative about the optimal action in the states that follow temporally while providing information about the optimal action in faraway states. Thus, such local structure represents a nontrivial assumption about the kinds of problems we are interested in.

\section{Option Iteration}
Rather than learning a single strong policy, OptIt aims to learn a set of policies such that, in every state we encounter, at least one policy in the set is good for some horizon into the future. To learn such a set, we adopt an approach similar to ExIt by optimizing for agreement between the learned set of option policies and the results of a relatively computationally expensive search. In particular, we maintain a set of $N$ option policies for $n\in\{0,...,N-1\}$, we denote the probability of sampling action $a$ in state $s$ with option $n$ as $\pi_n(a|s; \theta)$.\footnote{Throughout we use $\theta$ generically to represent the complete set of learnable parameters defining an agent, though in general only a subset will be used by each function.}

For each contiguous trajectory segment of $K$ steps, we aim to have a weighted mixture of the option policies in our learned set which is a close match to the results of a search procedure jointly for all $K$ steps. 
More precisely, consider a given trajectory segment $s_0,s_1,...,s_{K-1}$ along with $\tilde{\pi}(\cdot|s_k)$ representing the search policy, that is, the improved policy returned by running a search procedure in $s_k$. We optimize the following loss for randomly sampled segments of $K$ trajectory steps from a replay buffer
\begin{multline}
    \mathcal{L}=\E\limits_{A_k\sim\tilde{\pi}(A_k|s_k)}\Biggl[\\-\log\left(\sum_{n=0}^{N-1}\rho(n|s_0;\theta)\prod_{k=0}^{K-1}\pi_n(A_k|s_k;\theta)\right)\Biggr],
\end{multline}\label{option_distillation_loss}
where $\rho(n|s_0;\theta)$ is a learned policy over options conditioned on only the first state in the sequence.
Using a weighting conditional on only $s_0$ reflects the fact that, when planning, we must select an option conditional on only $s_0$. If instead $\rho$ was conditioned on the whole trajectory segment, we may end up with one of two different options being best depending on random transitions following $s_0$. This is undesirable given we wish to use the options for planning from the current state and thus want some option to be good in expectation conditional on the current state, rather than only conditional on some random future states. Equation~\ref{option_distillation_loss} is exactly the cross-entropy between the search policy for each state in the sequence and the joint distribution over action induced by the weighted mixture of option policies.

Note that $A_k$ is sampled independently from the action actually executed in the environment. In practice, we stochastically sample $A_k\sim\tilde{\pi}(A_k|s_k)$ in each update rather than optimizing the expectation in Equation~\ref{option_distillation_loss} directly as the latter is intractable for moderately large $k$.

Even if $\rho$ is uniform over options, Equation~\ref{option_distillation_loss} reduces to the expected LogSumExp over the log-likelihood of the search action sequence under each option. LogSumExp acts as a smoothed maximization, putting more emphasis on higher-value elements. Thus, relative to simply optimizing the average log-likelihood over options, Equation~\ref{option_distillation_loss} will tend to favour making options that are already a good match better. The emphasis on good options will be even stronger if $\rho$ is well learned. This motivates the intuition that OptIt aims to learn a set of options such that, for every state we encounter, at least one is good for some horizon into the future.

Note the importance of optimizing for agreement over $K$ steps rather than just a single step. If we optimize for single-step agreement there is unlikely to be a benefit to using a set of options instead of a single policy as the weighted mixture itself would collapse to a policy over actions.

It is useful to contrast OptIt with ExIt. Instead of trajectory segments, ExIt considers individual states. Instead of a set of policies, ExIt optimizes a single policy to agree with the search results in all states by optimizing the following loss on states sampled from a replay buffer:
\begin{equation*}
\mathcal{L}=\E\limits_{A\sim\tilde{\pi}(A|s)}\left[-\log(\pi(A|s;\theta))\right].
\end{equation*}

\newpage
\subsection{Option Iteration with Monte Carlo Search}
\begin{algorithm}[tb]
   \caption{MCS with Options}
   \label{MCS_with_options}
\begin{algorithmic}
   \STATE \textbf{Input:} $\theta, \bar{\sigma}, s$
   \FOR{$a$ in $\mathcal{A}$}
   \FOR{$n$ in $0:N-1$}
   \STATE $M = simulation\_budget/(|\mathcal{A}|N)$
   \STATE $\hat{Q}_n(a,s)\leftarrow0$
   \FOR{$j$ in $0:M-1$ (in Parallel)}
   \STATE Sample $s_1\sim p(\cdot|s,a)$
   \STATE Simulate $\pi_n(\cdot|\cdot;\theta)$ for $K-1$ steps from $s_1$
   \STATE Get simulated trajectory $(s_0,a_0,r_1,...,s_K)$
   \STATE $G\leftarrow\sum\limits_{k=0}^{K-1}\gamma^kr_{k+1}+\gamma^Kv(s_K;\theta)$
   \STATE $\hat{Q}_n(a,s)\pluseq G/M$
   \ENDFOR
   \ENDFOR
   \ENDFOR
   \STATE $\tilde{p}(a,n)\propto\exp(\hat{Q}_n(a,s)/(\bar{\sigma}\beta))$
   \STATE $\tilde{\pi}(a|s)\leftarrow\sum\limits_n \tilde{p}(a,n)$
   \STATE $\tilde{a}\leftarrow \argmax\limits_a \max\limits_n \hat{Q}_n(a,s)$
   \STATE $\tilde{v}\leftarrow \max\limits_{a,n}\hat{Q}_n(a,s)$
   \STATE \textbf{return} $\tilde{a}, \tilde{v}, \tilde{\pi}$
\end{algorithmic}
\end{algorithm}

OptIt could be integrated with any planning algorithm that uses search to compute an improved policy in each visited state, including Monte-Carlo Tree Search (\citet{kocsis2006bandit,coulom2006efficient}; MCTS). However, Monte-Carlo Search (\citet{tesauro1996line}; MCS) provides a particularly appealing use case as the addition of learned options could help to mitigate some of its inherent weaknesses. Unlike MCTS, MCS does not perform policy improvement in non-root nodes during the search, each action is only evaluated under a specific rollout policy. On the other hand, MCS offers certain advantages relative to MCTS. In particular, MCS is straightforward to parallelize and also straightforward to apply to stochastic environments.

Performing MCS with a strong learned set of options, can help to mitigate the drawbacks and allow us to capitalize on the benefits inherent in its simplicity. By searching in a joint space of actions and options rather than just primitive actions, we can evaluate each action under various possible behaviours rather than a single learned rollout policy. 
Unless one policy in the set is best everywhere, finding the best action under the best policy in a set for each encountered state will give us better actions overall than selecting the best action under any single policy from the set. If the set of learned options is sufficiently rich to capture the plausible future behaviours then it may not be necessary to explicitly build a tree over future trajectories.

We will apply OptIt on top of MCS in our experiments leaving its application to other search algorithms for future work.

We run MCS in the joint space of options and initial actions. Effectively, we search for the best combination of initial action and subsequent behaviour from the set of available learned options. MCS with any finite set of options and actions effectively reduces the search problem to a finite-armed bandit setting. We could thus choose any finite-armed bandit algorithm to select options and actions during search. To avoid complex interactions between the bandit algorithm and policy learning procedure, which could confound our main focus, we make a simple choice in this work. Namely, we fix the total number of simulations used in each state and allocate this simulation budget evenly amongst the joint space of options and initial actions. For example, if we allocate $1000$ total simulations with 5 options and 4 actions, then for each option $n$ and action $a$ we would perform $1000/(4\cdot5)=50$ simulations where initial action $a$ is selected and the policy of option $n$ is followed thereafter for the remainder of the rollout length. Applying OptIt with more sophisticated choices of bandit algorithm such as PUCB~\citep{rosin2011multi} or Sequential Halving~\citep{karnin2013almost, danihelka2021policy} is an interesting direction for future work.

While we run MCS in the joint space of options and actions, we ultimately use the search results only to select an action to execute and provide an improved policy over actions alone as a target for option learning. To do this, we first compute a joint distribution over options and actions based on the entropy regularized average return resulting from the simulations:
\begin{equation*}
\tilde{p}(a,n)\propto\exp(\hat{Q}_n(a,s)/(\bar{\sigma}\beta)),
\end{equation*}
where $\hat{Q}_n(a,s)$ is the average $K$ step bootstrapped return for simulations where the initial action is $a$ and option $n$ is followed thereafter, $\beta$ is an entropy regularization hyperparameter, and $\bar{\sigma}^2$ is an exponentially weighted average of the variance of returns for recent rollouts used to reduce sensitivity to $\beta$ across problems. We next compute the search policy over actions by marginalizing out the options $\tilde{\pi}(a|s)=\sum_n \tilde{p}(a,n)$, the action $\tilde{a}$ returned by the search to actually execute in the environment is $\tilde{a}=\argmax\limits_a \max\limits_n \hat{Q}_n(a,s)$. We also learn a value estimate, which is used only in computing bootstrapped returns during search, the target for which is simply $\tilde{v}=\max\limits_{a,n}\hat{Q}_n(a,s)$. 

Pseudocode for our implementation of MCS with options is displayed in Algorithm~\ref{MCS_with_options}. Since we consider episodic environments, we define $v(\bot;\theta)=0$ for the terminal state.

After searching in each state, the resulting $\tilde{\pi}$ and $\tilde{v}$ are stored in a replay buffer along with the current state $s$. The option policies and the policy over option $\rho$ are trained using Equation~\ref{option_distillation_loss} on randomly sampled batches of contiguous length $K$ trajectory segments.\footnote{We sample start states randomly and truncate segments at terminal or timeout states. We also divide the loss by the trajectory segment length, only counting truncated steps in case of truncation.} The value function is trained to minimize squared error relative to $\tilde{v}$.

\section{Does Option Iteration Learn Useful Options for Monte-Carlo Search?}
Here, we evaluate the ability of OptIt to discover useful options from the results of search. In all our experiments we use simple feed-forward neural networks for function approximation, with binary observations as input. Option policies as well as the policy over options $\rho$ are implemented using a network with a single shared trunk with only the output layer differing, hence the difference in parameter count due to using multiple options versus a single policy is very minor. The value function is approximated using a separate feed-forward network of equal size. We begin with a simple domain as a pedagogical illustration before moving on to a more challenging and plausible planning problem. 
\subsection{An Illustrative Domain Highlighting the Benefit of Option Iteration}\label{sec:compass}
\begin{figure}[tb]
    \centering
    \includegraphics[width=0.8\columnwidth]{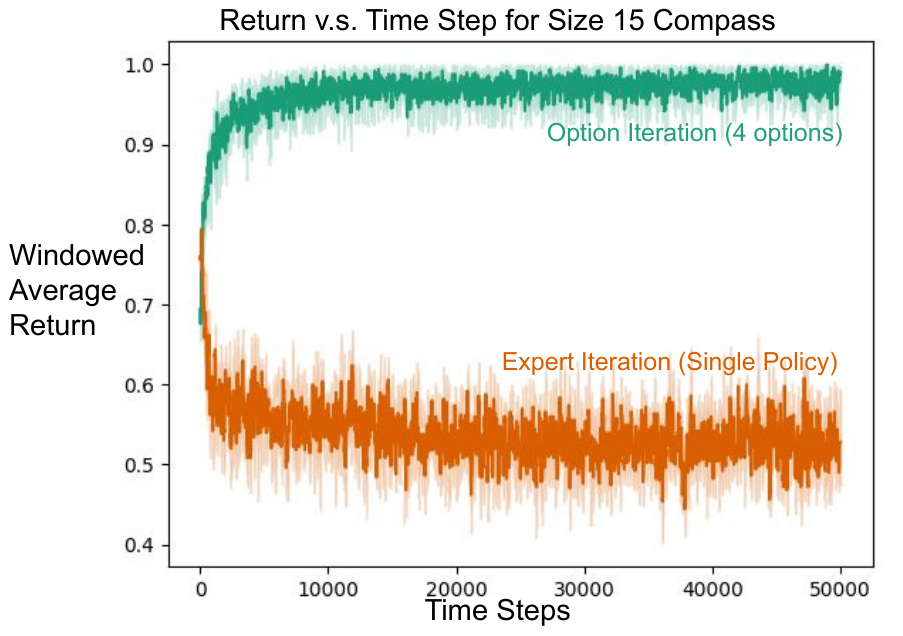}
    \caption{Windowed average return over training time for Monte-Carlo search with options learned via OptIt compared to Mote-Carlo Search with just primitive actions using a rollout policy learned via ExIt. Here, and in all other results, we report undiscounted return up to either termination or timeout. Error bars show 95\% confidence interval over 5 random seeds.}
    \label{Compass_results}
\end{figure}
We first illustrate the benefit of OptIt in a variant of the Compass domain, as described in Section~\ref{joint_action_distribution}. The domain used here differs slightly from that described in Section~\ref{joint_action_distribution} in that we do not provide the index $m$ identifying the grid. Instead, there is simply a reward of $1$ on one edge and a reward of $-1$ on the other edges at random. Which edge the positive reward is located on is not observable and can be determined by the agent only by performing rollouts under the model. This is essentially analogous to the case where $M$ is so large that it is unlikely that the agent will ever see the same grid twice.

We note that this situation is perhaps somewhat contrived in that information about which edge the reward is on is available to the world model, but impossible to determine directly from the agent's observations. However, it serves as a straightforward empirical verification of the arguments in favour of OptIt laid out in Section~\ref{joint_action_distribution}. We further argue that this is a reasonable surrogate for the more plausible situation where the optimal policy is a function of the agent's observations, but in a nuanced way that has not been well learned due to either limited policy training or policy network capacity, which can nonetheless be determined by simulation using the world model. Subsequent sections will evaluate OptIt in environments without such contrived partial observability.

We set the width of the grid to 15. The timeout period for an episode was set to 20 such that there was enough time to reach the positive reward from any initialization position under the optimal policy. After the time-out period a new episode begins regardless of whether the terminal state has been reached.\footnote{Note that timeouts are not applied during search rollouts.} The rollout length for options and length $K$ of trajectory segments used in the OptIt loss are likewise set to 20 such that it is possible to reach any edge of the grid in one rollout. In this experiment, we use a relatively low simulation budget of 50 rollouts. We test an agent using 4 options learned with OptIt against an agent operating in the space of primitive actions with a rollout policy learned from search results using ExIt. Other hyperparameter settings are available in Appendix~\ref{hyperparameters}.

The results are displayed in Figure~\ref{Compass_results}. OptIt is able to quickly learn good approximations to the four directional options that allow the search to rapidly locate the positive reward on any of the 4 edges and from that point forward achieves near-perfect performance. On the other hand, the single rollout policy learned using ExIt is high entropy and chaotic, albeit with an inconsistent tendency to head toward the nearest edge, making it rare to reach the correct edge of the grid in 20 steps. Visualizations of the learned options and policy are provided in Appendix~\ref{compass_options}.

\subsection{Evaluating Option Iteration in a Challenging Planning Domain}\label{sec:EPM}
\begin{figure}[tb]
    \centering
    \includegraphics[width=0.8\columnwidth]{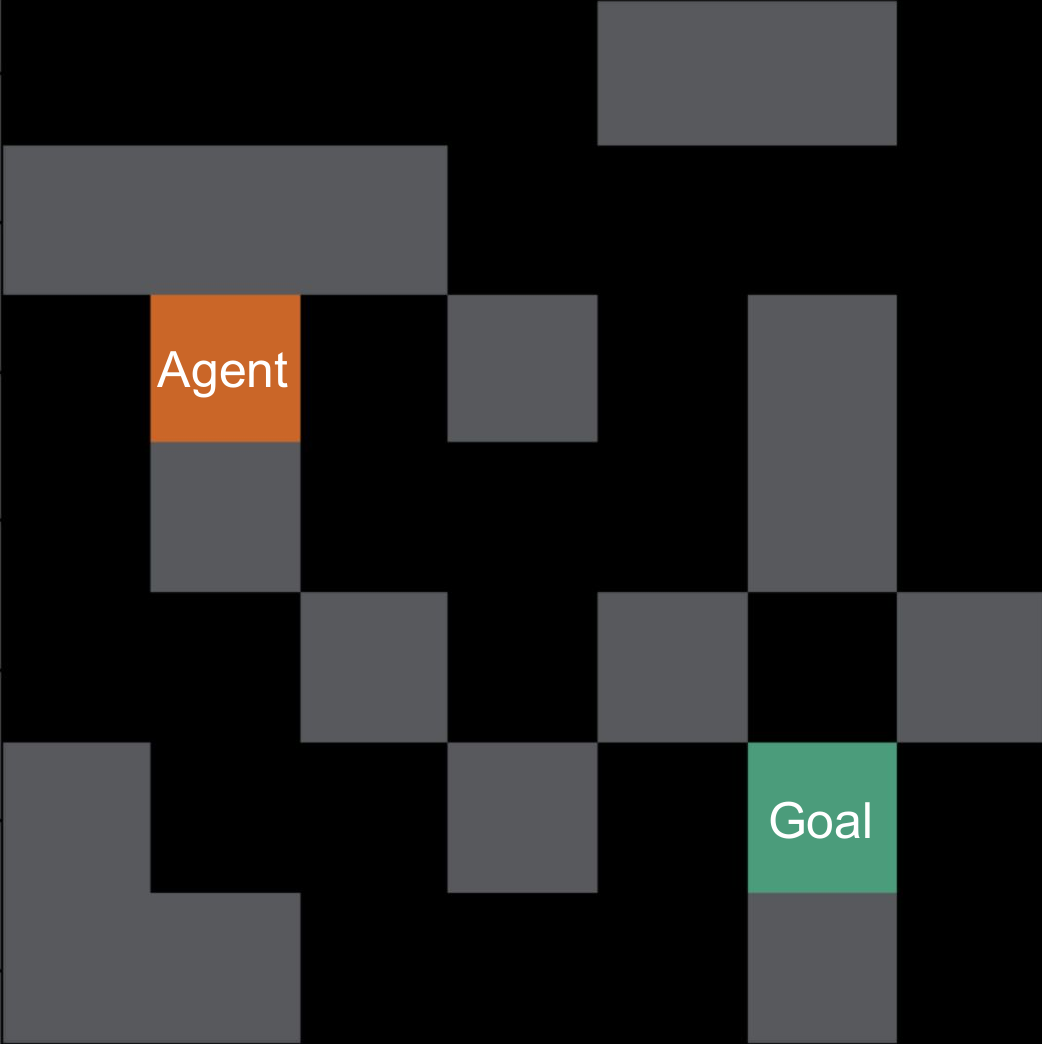}
    \caption{Example of a state in the 7x7 ProcMaze environment. Note that the maze layout is randomized in each episode, making the state space much larger than that of a particular maze.}
    \label{procmaze_env}
\end{figure}
We now empirically evaluate OptIt on a conceptually simple environment designed to present a significant planning challenge. Specifically, we use a variant of the ProcMaze environment of~\citet{young2022benefits}, that we refer to as ElectricProcMaze for reasons we will explain shortly.

Unlike the environment in Section~\ref{sec:compass} ElectricProcMaze is fully observable. There no information a priori hidden from the policy network that affects the correct behaviour which would require the use of options to represent the joint distribution of optimal actions. However, we expect this environment is challenging enough that learning could benefit from representing a distribution of possible behaviours before the single optimal policy is well learned.

ProcMaze consists of procedurally generated gridworld mazes which require navigating from a start state to a goal state. The maze layout, start state and goal state are randomized at the start of each episode. A reward of -1 is given for each step until the goal is reached, at which point the episode terminates.
The observations consist of a flat binary vector including one hot vectors for the goal location and agent location, and vectors indicating the presence or absence of a wall. 
The action space consists of moving in each cardinal direction.
A new maze is generated in each episode using randomized depth-first search. 
An example of a possible state is displayed in Figure~\ref{procmaze_env}. 

In ElectricProcMaze, instead of remaining in place upon transitioning into a wall, the agent is allowed to move into the wall cell but with a large negative reward (analogous to an electric shock in animal experiments) set to be equal to one more than the largest possible number of steps required to reach the goal across all possible maze configurations.

We use ElectricProcMaze rather than ProcMaze as the latter has the property that the greedy policy with respect to the action value function of the random policy is optimal while ElectricProcMaze does not. \citet{laidlaw2023bridging} demonstrated that this characteristic is surprisingly common, particularly in environments where standard deep reinforcement learning algorithms perform well. However, it is undesirable for the present work where the main focus is on policy learning for search. See Appendix~\ref{EPM_motivation} for further details on this point.

We present experiments in 7x7 mazes. We found in preliminary experiments that smaller mazes were easily solved by MCS in the space of primitive actions and hence presented little potential to benefit from options. We also use a timeout of 120 steps after which the episode is ended to avoid the agent getting stuck indefinitely on a single challenging maze.

We evaluate OptIt with MCS using 5 learned options against ExIt with a single learned policy.
ExIt is implemented using the exact same code as OptIt, simply reducing the number of options to 1.\footnote{This means that ExIt is also trained on sequences and uses samples from the search policy rather than directly minimizing cross-entropy. We also tried implementing ExIt with independent samples and exact cross-entropy but surprisingly found this to be significantly worse, hence we report results with the same setting as OptIt for simplicity. We include an ablation study to better understand these choices in Appendix~\ref{sec:ExIt_ablation}.} In order to establish a strong performance baseline for ExIt on this environment we sweep the step-size and entropy regularization parameters and select the one with the best average return in the final 100,000 of 500,000 training steps, see Appendix~\ref{hyperparameters} for details. We then use the same hyperparameters for OptIt. Each is given the same simulation budget per step, which for OptIt is distributed evenly over all option-action combinations and for ExIt is distributed over only actions.
In each case, the value function is used to estimate the value of the final state in length 5 rollouts during MCS. As an additional baseline, we also tested training 5 options by optimizing the mean cross-entropy over options, that is
\begin{equation*}
        \mathcal{L}=\E\limits_{A_k\sim\tilde{\pi}(A_k|s_k)}\Biggl[-\frac{1}{N}\sum_{n=0}^{N-1}\log\left(\prod_{k=0}^{K-1}\pi_n(A_k|s_k;\theta)\right)\Biggr].
\end{equation*}

\begin{figure}[tb]
  \centering
  \includegraphics[width=\columnwidth]{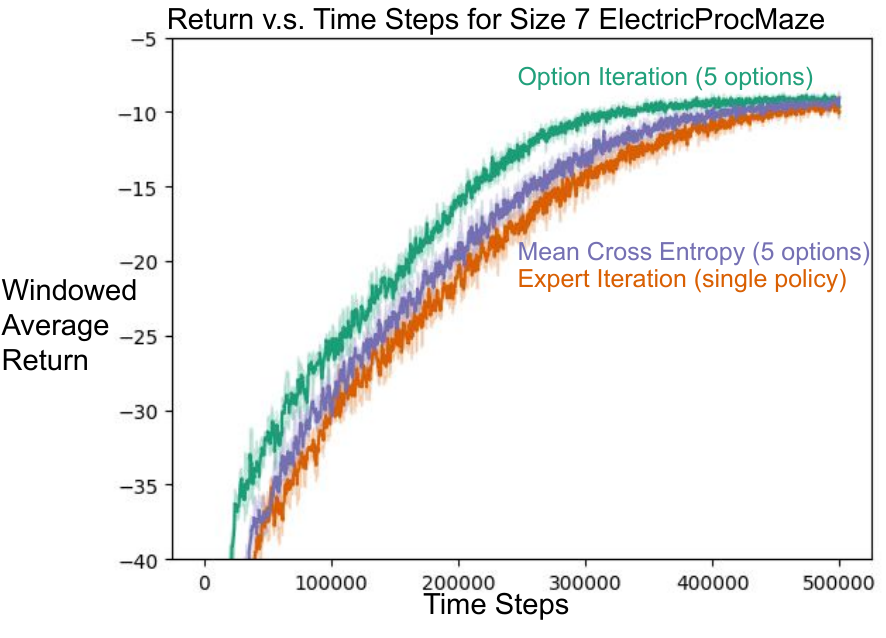}
   \caption{Windowed average return over training time for OptIt and baselines on size 7 ElectricProcMaze. Error bars show 95\% confidence interval over 5 random seeds. The y-axis is thresholded at $-40$ to omit the rapid period of initial improvement. Running MCS in the space of options learned with OptIt provides a significant benefit compared to using primitive actions evaluated under the single learned rollout policy. 
   }\label{ProcMaze_results}
\end{figure}

The results are displayed in Figure~\ref{ProcMaze_results}. Planning with options discovered by OptIt shows a clear benefit over using a single learned rollout policy. We also see that using the OptIt loss significantly outperforms the mean-cross entropy baseline. The latter performs just slightly better than ExIt, suggesting that the specialization induced by maximizing the joint likelihood of action sequences results in a more useful set of options.

\subsection{Evaluating Option Iteration on a Challenging Planning Domain with Inherent Hierarchical Structure}
In Section~\ref{sec:EPM} we demonstrated that using options learned by OptIt in MCS can significantly accelerate learning in a challenging planning domain compared to using a single policy learned by ExIt. This is despite the fact that randomized mazes have no obvious hierarchical structure. We may expect to see more significant benefits in domains where there is some underlying hierarchical structure in the environment. In this section, we introduce another conceptually simple environment, built on top of ElectricProcMaze,  to test this.

We call the hierarchical environment introduced here HierarchicalElectricProcMaze. This environment consists of a base-environment (in this case ElectricProcMaze) and a controller-environment. The basic idea is to create a simplified abstraction of an agent interacting with a low level controller such that executing a single action in the base-environment via the controller requires a sequence of coherent action in the controller-environment. A more complex example would be a robot learning to play chess when it is required to physically manipulate the pieces to take actions in the game. 

The controller-environment consists of an $8\times 8$ grid with buttons placed in the centre of each edge. Each button corresponds to a particular action in the base-environment. The action space consists of moving in the 4 cardinal directions, and in this case will move the agent in the controller-environment. Upon reaching a button in the controller-environment the selected action to execute in the base-environment will be changed to the action associated with the button. On every 8th time-steps in the controller-environment, the currently selected action will be executed in the base-environment which will be advanced by one time-step. At this time, the selected action will be reset to no-op. Executing no-op will cause the agent not to move in the base-environment, hence to select an action besides no-op in the next base-environment step, the agent must once again move to a button within 8-steps. The environment is fully observable with the observation space consisting of the base-environment observations, together with one-hot vectors indicating the agent's position in the controller-environment, the currently selected base-environment action, and the remaining time-steps until the next base-environment action is executed. Rewards are simply the rewards from the base environment and fixed to zero except on time-steps when the base-environment is advanced.

We use size 5 ElectricProcMaze as the base-environment, reduced from size 7 in Section~\ref{sec:EPM} to compensate for the added difficulty of having to control the environment indirectly. We maintain most of the hyperparameters from Section~\ref{sec:EPM}. We changed the option rollout length from 5 to 8 to match the size of the controller-environment grid, corresponding to one base-environment action per rollout. We also increase the capacity of the network as the original network worked poorly for both OptIt and ExIt in preliminary experiments indicating significant underparameterization. 

For HierarchicalElectricProcMaze, we also reduced the entropy regularization factor $\beta$ from $0.1$ to $0.01$ for Option Iteration which we found was necessary to obtain a performance benefit compared to ExIt. In light of this change, to facilitate a fair comparison, and a more complete picture of the behaviour of each approach, we performed a sweep over $\beta$ values in powers of 10 and display the performance of each algorithm for the best $\beta$. The results of the sweep are available in Figure~\ref{ShipProcMaze_results} in Appendix~\ref{hyperparameters}. The results of this sweep reveal that OptIt achieves optimal performance at an order of magnitude lower level of entropy regularization than ExIt and remains more robust to further reduction in the regularization. One plausible interpretation of this result is that since OptIt is effectively learning a distribution of possible behaviours rather than fitting a single policy, it is able to maintain adequate behavioural diversity while fitting to a sharper search policy. This may in turn allow it to benefit from extracting more information from the search results without suffering as much from overfitting.

\begin{figure}[tb]
  \centering
  \includegraphics[width=\columnwidth]{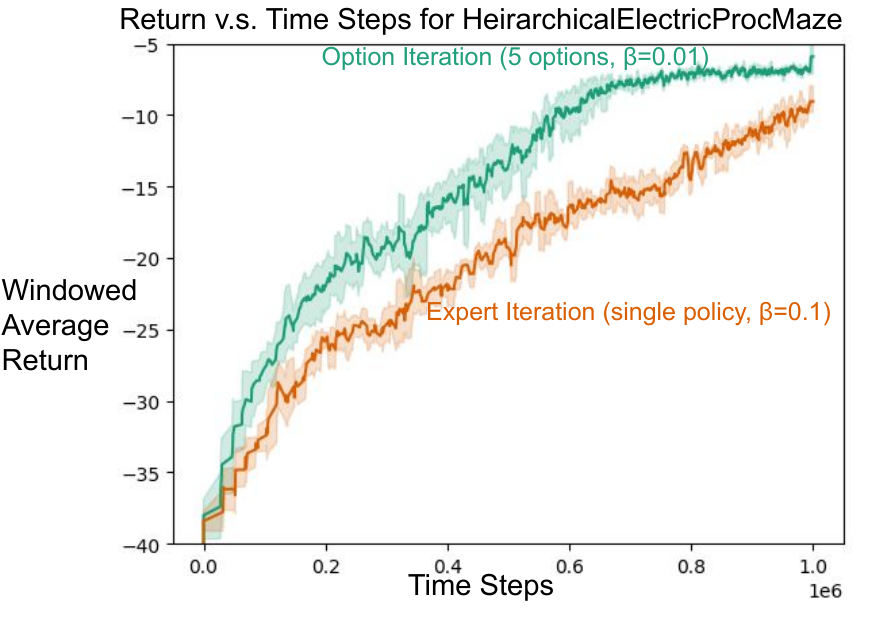}
   \caption{Windowed average return over training time for OptIt and baselines on HierarchicalElectricProcMaze with size 8 controller and size 5 base-environment. Each algorithm is presented for its best $\beta$ value from a sweep in powers of 10. Error bars show 95\% confidence interval over 5 random seeds. The y-axis is thresholded at $-40$ to omit the rapid period of initial improvement. Running MCS in the space of options learned with OptIt provides a large benefit compared to using primitive actions evaluated under the single learned rollout policy.
   }\label{ShipProcMaze_results}
\end{figure}
The results are displayed in Figure~\ref{ShipProcMaze_results}. Planning with options discovered by OptIt shows a large benefit over using a single learned rollout policy.

Unlike the options learned for the Compass environment in Section~\ref{sec:compass}, we did not observe the options learned for HierarchicalElectricProcMaze to be particularly interpretable. Nevertheless, we did find the learned options to display significant behavioural diversity in an analysis which we present in Appendix~\ref{ship_EPM5_options}.

\section{Related Work}
Our work builds on the options framework of~\citet{sutton1999between}.  In particular, we focus on the problem of option discovery, where rather than providing an agent with a set of prespecified options, we wish to design algorithms that allow agents to build a set of useful options on their own. 

A variety of methods have been proposed for option discovery, with a number of different motivations behind them. Some approaches aim to directly optimize options to facilitate good performance on a task or distribution of tasks~\citep{bacon2017option, frans2017meta, veeriah2021discovery}. Others aim to learn options which help navigate between disparate regions of state-space, for example by identifying bottleneck states~\citep{mcgovern2001automatic,stolle2002learning}, exploiting graph-theoretic properties of the transition dynamics~\citep{machado2017laplacian, klissarov2023deep}, or encouraging options to contain a lot of information about their state at termination~\citep{gregor2016variational, eysenbach2018diversity, harutyunyan2019termination}. A large body of work considers learning hierarchical policies in which a high-level policy, trained to maximize task performance, outputs subgoals that a low-level policy is trained to achieve~\citep{dayan1992feudal, vezhnevets2017feudal, hafner2022deep}. 

\citet{jinnai2019finding} and \citet{wan2022toward} share our focus on discovering options that facilitate planning, albeit using value iteration rather than decision-time planning. \citet{co2018self} jointly learn a latent conditioned policy and trajectory level model such that the model predicts the trajectory resulting from the policy when conditioned on a particular latent state. The resulting latent state and model are then used in a decision-time planning procedure, similar to the way we use our learned set of options.

There is also a body of work focusing on discovering sets of options which fit a dataset of unsegmented expert demonstrations~\citep{zhu2022bottom, shankar2020learning, kipf2019compile, fox2017multi, krishnan2017ddco}. The present work can be seen as an instance of this where the demonstration data is dynamically generated by a search procedure and the discovered options are used to improve the search.

The present work focuses on option discovery in a single-task setting, while much of the literature focuses on the multi-task setting. The distinction is not always clear-cut but to give some examples: \citet{frans2017meta}, \citet{veeriah2021discovery}, and \citet{wan2022toward} all expressly focus on finding options which are useful across a distribution of related tasks as opposed to one specific task. In a sense, the multi-task setting is more natural for option discovery as one can imagine learning a set of options which capture temporal regularities in the optimal policy shared across the tasks. For a single task, there is necessarily a single optimal policy, making it less obvious why we'd want to learn a set of behaviours. In Section~\ref{joint_action_distribution}, we motivate the benefit of options in a single, complex, task by the desire to quantify uncertainty in the joint distribution over actions. One can think of this as being loosely related to the multitask setting. We have replaced the assumption of multiple ground truth tasks, each with a distinct optimal policy, with a distribution over optimal policies induced by uncertainty, despite there being only a single ground truth task.

In Algorithm~\ref{MCS_with_options}, we compute an improved policy and value function based on Monte-Carlo evaluation under a set of possible options. This is an instance of Generalized Policy Improvement, introduced by~\citet{barreto2017successor}.

As already discussed, OptIt is inspired by the ExIt ~\citep{anthony2017thinking} and the closely related approach used in AlphaZero~\citep{silver2017mastering}. Recently,~\cite{zahavy2023diversifying} have demonstrated that optimizing a diverse set of players, encoded as different latent state inputs to a single network, using a quality-diversity objective can significantly improve the performance and robustness of AlphaZero.

\section{Conclusion and Future Work}
We have introduced Option Iteration, a simple approach to discovering options by amortizing the results of a computationally expensive search algorithm. In turn, the discovered options can be used to improve future searches, resulting in a virtuous cycle where better options lead to improved search, which in turn enables the discovery of better options. OptIt is motivated by the desire to capture information about the joint distribution over near-optimal actions in a trajectory that it is not possible to capture with a single state-conditioned policy. We have demonstrated empirically that the options learned by OptIt provide a significant benefit over a single policy learned with ExIt when planning with MCS in challenging planning domains.

There are a number of interesting directions for future work on OptIt, we highlight a few below:
\begin{itemize}
\item Rather than fixing the horizon of the learned options one could learn termination functions for each option based by maximizing the likelihood of trajectories under the distribution induced by the combination of the option-policies, policy over options and termination functions. In this case, the termination functions would aim to learn the optimal place to split the trajectory into segments such that in each segment the search policy was closely matched to a particular option. We discuss how this could be done in detail in Appendix~\ref{learned_option_termination}, building on the approach of~\citet{fox2017multi}.
\item One could explore applying OptIt in combination with other search algorithms such as MCTS. 
\item Rather than using a discrete set of options, one could specify options with an arbitrary latent variable. 
\item OptIt inherently relies on the reward signal to allow the search procedure to identify good actions. This is limiting in domains with very sparse rewards, in which we might instead want to discover options that are useful for exploration even before the agent has managed to locate any reward. One simple way to address this would be to run OptIt in combination with an intrinsic reward signal.
\item In this work, we encourage option diversity only implicitly by maximizing trajectory likelihood under the $\rho$ weighted mixture of option policies. It may be beneficial to explicitly encourage diversity, for example using a quality-diversity approach similar to that of~\citet{zahavy2023diversifying}. 
\end{itemize}

\subsubsection*{Acknowledgments}
We thank Michael Bowling, Martin Mueller, Aditya Ramesh, Louis Kirsch, Yi Wan, Arsalan Sharifnassab, and Tian Tian for useful conversations. Funding for this work was provided by NSERC, Alberta Innovates, DeepMind, CIFAR and Amii. Computational resources for this work were provided by the Digital Research Alliance of Canada.

\bibliography{paper}
\bibliographystyle{icml2023}

\newpage
\appendix
\onecolumn
\section{Motivation for Introducing ElectricProcMaze}\label{EPM_motivation}
\begin{figure}[htb]
    \centering
    \includegraphics[width=0.25\columnwidth]{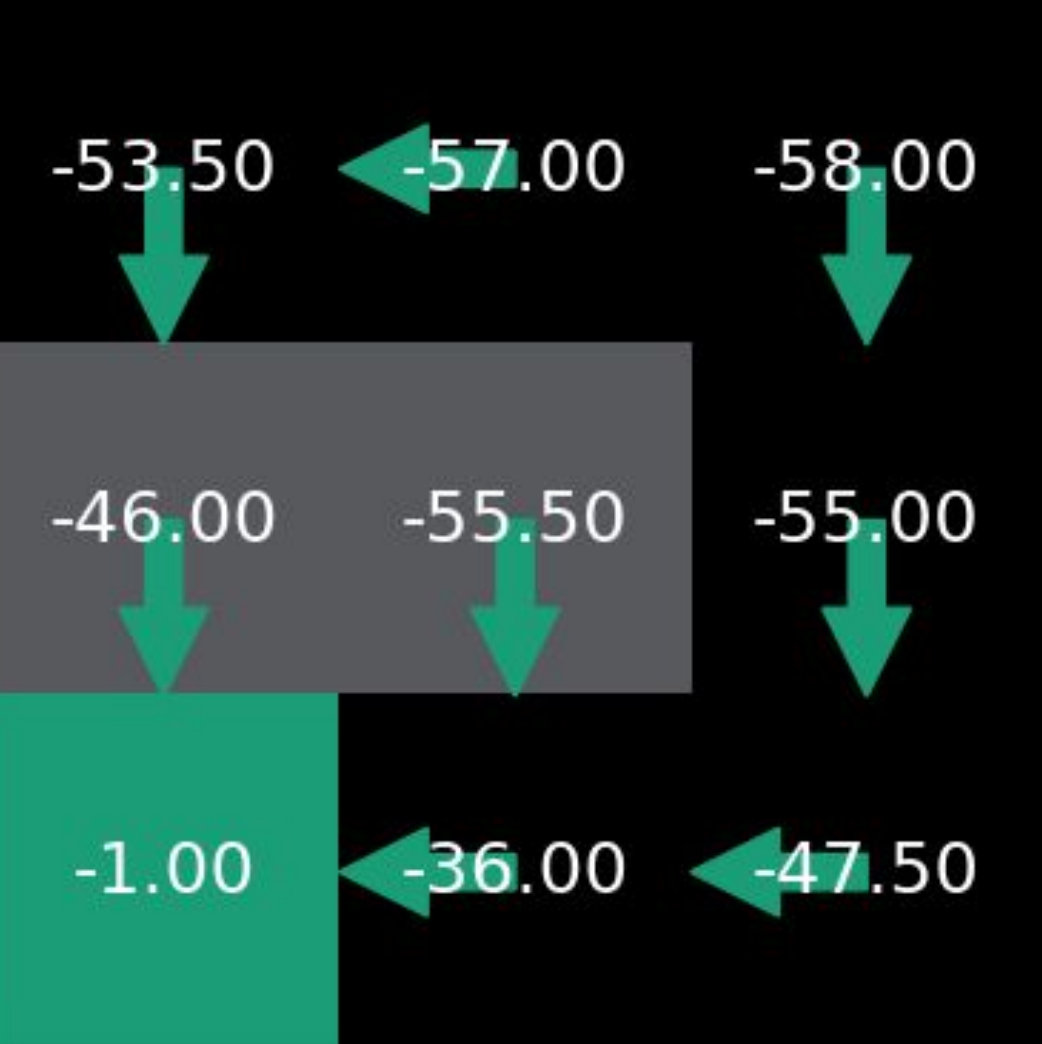}
    \caption{A simple ElectricProcMaze instance showing the greedy policy (green arrows) and associated action values of the random policy. The values printed on each cell are the action-values associated with moving into that cell from any other cell which works in this case since the reward is determined by the cell being entered regardless of the origin. The action values were computed by explicitly solving the Bellman equation with a uniform random behaviour policy. The reward for entering a wall cell is $-7$ in this case which is one less than the return achievable by following an optimal policy from the furthest cell to the goal. Hence the optimal policy will never enter a wall cell, but the greedy policy with respect to the action values of the random policy does.}
    \label{EPMGreedy}
\end{figure}
ElectricProcMaze is a variant of the ProcMaze environment from~\cite{young2022benefits} where instead of remaining in place upon transitioning into a wall, the agent is allowed to move into the wall cell but with a large negative reward (analogous to an electric shock in animal experiments) set to be equal to one more than the largest possible number of steps required to reach the goal across all possible maze configurations.

Our motivation for using ElectricProcMaze rather than ProcMaze stems from observations we made in preliminary experiments on ProcMaze. In particular, we observed that using a random rollout policy (and thus relying on only value function learning to guide the search) was almost as effective as learning a single rollout policy in this case. This can be explained by noticing that due to there being only one path to the goal, ProcMaze has the characteristic that the greedy policy with respect to the action value function of the random policy is optimal. \citet{laidlaw2023bridging} have demonstrated that this characteristic is surprisingly common, particularly in environments where standard deep reinforcement learning algorithms perform well. This does not necessarily mean that policy learning cannot be beneficial, as the number of samples required to evaluate the random policy to sufficient accuracy to repliably identify the greedy action may be much larger than the number of samples required under an improved policy. It does however mean that policy learning is relegated to a more minor role then in the general case where it is necessary for convergence to the optimal policy when bootstrapping from multistep rollouts, regardless of the number of simulations used. This is undesirable when we are interested in investigating the benefits of learning multiple options in order to improve the quality of rollouts. 

ElectricProcMaze breaks the condition that the greedy policy with respect to the action value function of the random policy is optimal. The best action under the uniform random policy will often follow a shorter path to the goal which passes through walls, as doing otherwise will often mean hitting more walls in expectation over a random walk. This is demonstrated on a simple problem instance in Figure~\ref{EPMGreedy}. On the other hand, an optimal policy will always avoid walls by design since we set the penalty to be large enough that it is never better to move through a wall than to follow the open path.
\newpage
\section{Learned Options for Compass Environment}\label{compass_options}
\begin{figure}[htb]
    \centering
    \includegraphics[width=\columnwidth]{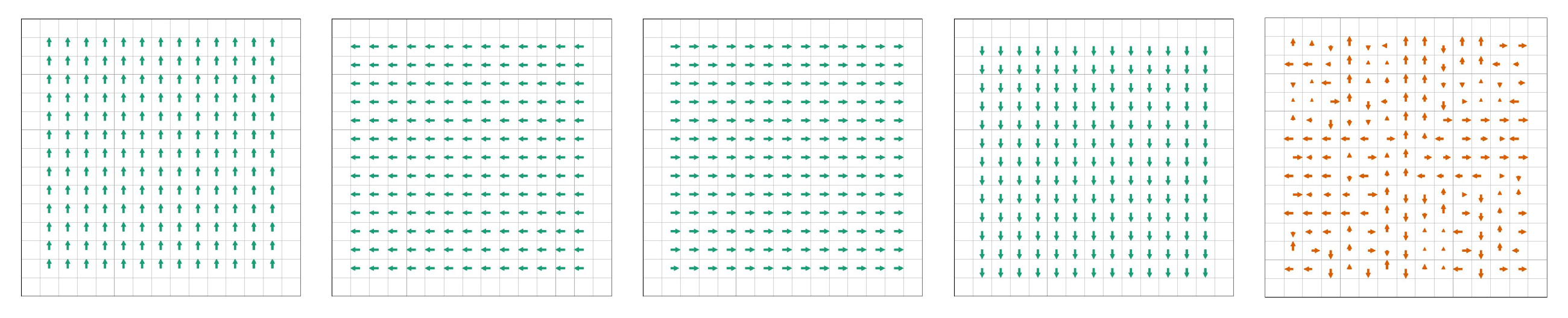}
    \caption{The left four plots with green arrows show the highest probability action selected by each of the 4 options learned by OptIt after 200,000 steps in the $15\times15$ compass world environment. The rightmost plot with orange arrows shows the action selected by the single policy learned with ExIt. In all cases the length of the arrow indicates the probability of the action with probability 1 corresponding to the arrow touching the edge of the grid cell. The options learned by OptIt correspond essentially perfectly to the 4 directional policies which reach the edges of the grid as quickly as possible. The single policy learned by ExIt is chaotic and has relatively high entropy in most grid cells.}
\end{figure}

\section{Expert Iteration Loss Ablation}\label{sec:ExIt_ablation}
\begin{figure}[htb]
  \centering
  \includegraphics[width=0.5\columnwidth]{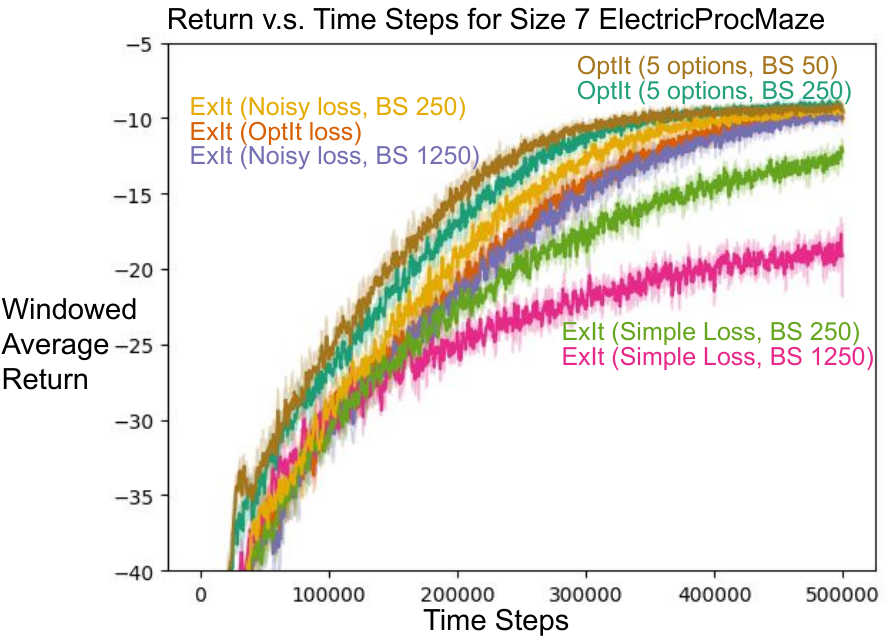}
   \caption{Windowed average return over training time for various loss function choices for ExIt and OptIt on size 7 ElectricProcMaze. OptIt loss refers to the same loss used to train OptIt which uses samples form the search policy and trains on length 5 sequences. Noisy Loss still uses samples from the search policy but trains on independent samples. Simple Loss trains on the full cross entropy with the search policy on independent samples. BS refers to the batch size used in the gradient update. Error bars show 95\% confidence interval over 5 random seeds. The y-axis is thresholded at $-40$ to omit the rapid period of initial improvement.
   }\label{ExIt_ablation}
\end{figure}
In this section, we explore the impact of the choice to optimize the single policy learned by ExIt using the same loss function as OptIt. In particular, this means we optimize ExIt on sequences rather than independent samples, though the latter would be more natural when learning a single policy. It also means we fit the learned policy to samples from the search policy in each update rather than directly minimizing cross-entropy loss. This was necessary for tractability when optimizing over joint cross-entropy for a set of options over a sequence, but not when optimizing a single policy. Note that these choices should not impact the expectation of the loss but do change the noise in the updates. To better understand the impact of these decisions, we present an ablation study evaluating ExIt trained with alternative loss functions. The results are displayed in Figure~\ref{ExIt_ablation}.

The first variant of ExIt we test directly minimizes cross-entropy on independent samples. Surprisingly, we found that this version was significantly worse than simply using the OptIt loss, to better understand why, we also tried a version that trains on individual samples instead of sequences but still fits to sampled actions from the search policy rather than directly minimizing the cross-entropy with the search policy in each update. This version recovers similar performance, indicating that the benefit was likely due to sampling the search policy actions as opposed to training on sequences. For each loss, we test ExIt using a batch size of 250 and a batch size of 1250, the latter matches the total number of samples\footnote{Modulo truncation due to episode ends.} per update to that of the sequence-trained variant with batch size 250 and option rollouts of length 5. We found that the smaller batch size learned faster in all cases. For completeness, we also tested OptIt with 5 options and a batch size of 50 (thus a total of 250 samples per batch when accounting for sequence length), this also learned faster than the original setting of OptIt suggesting that the original batch size was simply suboptimally large for both OptIt and ExIt.

The fact that fitting to samples from the search policy, rather than the expectation, improves performance is curious and may be of independent interest. Further exploration of this may be fruitful, but is beyond the scope of the present work. Throughout, we report results for ExIt and OptIt trained with the same loss, since using the sampled cross-entropy appears significantly beneficial and optimizing on sequences not detrimental for ExIt.

\section{Investigating the Learned Options for HierarchicalElectricProcMaze}\label{ship_EPM5_options}
Compared to the learned options for the Compass environment presented in Appendix~\ref{compass_options}, we did not find the options learned for HierarchicalElectricProcMaze to facilitate easy interpretation. Nevertheless, here, we make an effort to highlight some of the features of the learned options that may contribute to their improved search performance compared to using a single policy. We focus on the options learned at 500,000 training steps, in the middle of training before the performance has plateaued.

\begin{figure}[htb]
    \centering
    \includegraphics[width=\columnwidth]{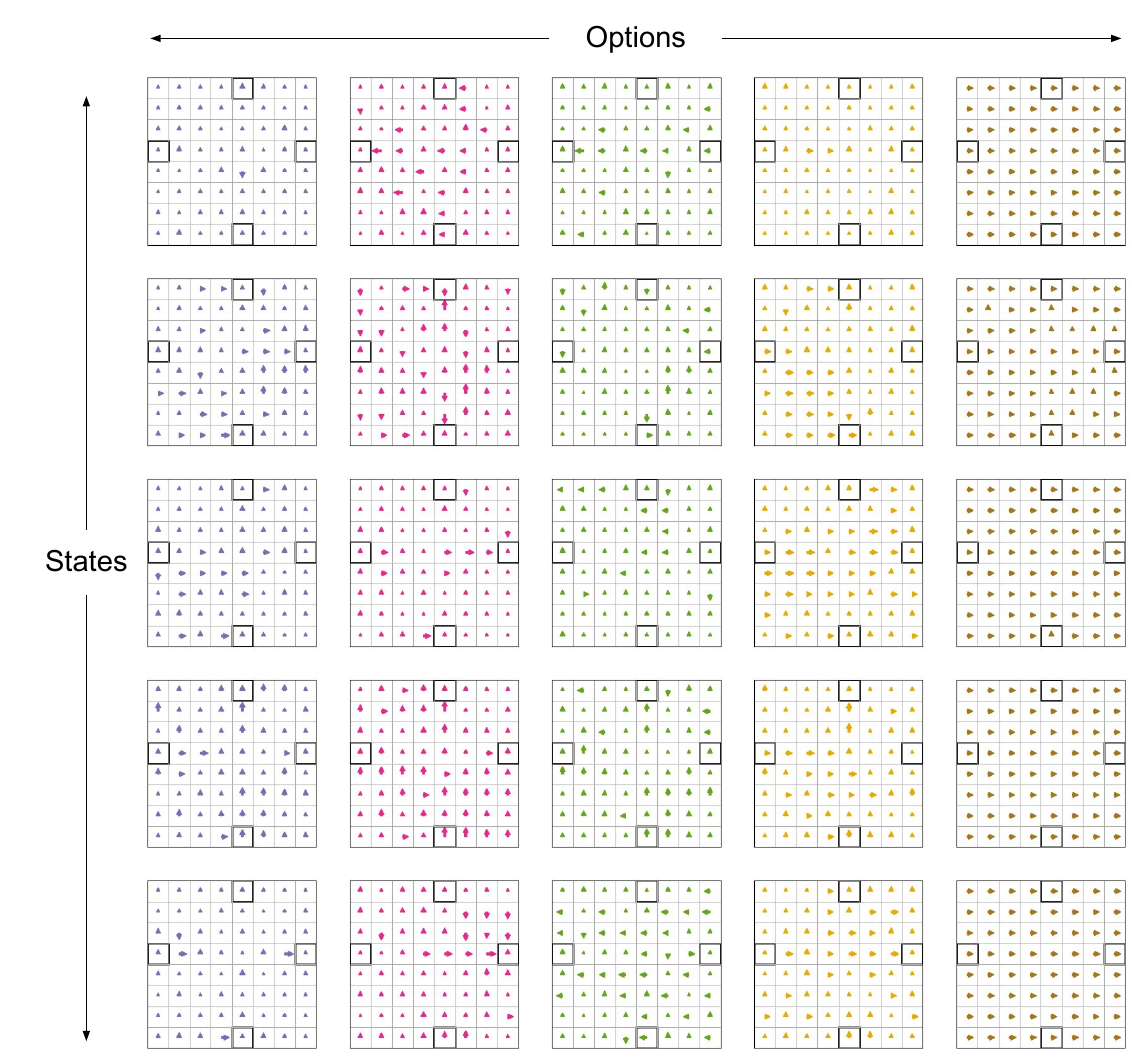}
    \caption{The 5 options learned by OptIt displayed for each state in the controller-environment grid across 5 different base-environment states in HierarchicalElectricProcMaze. The direction of each arrow shows the highest probability action in that cell, while the length of the arrow indicates the probability of that action with probability 1 corresponding to the arrow touching the edge of the grid cell. Button locations are highlighted with bold squares. Results are displayed for a single random seed corresponding to Seed 1 in Figure~\ref{option_button_frequency}.}
    \label{random_state_options}
\end{figure}

We first plot the behaviour of each of the 5 options across 5 random base-environment states while varying the controller environment state across the entire grid. The results are displayed in Figure~\ref{random_state_options}. The option in the far right column shows a strong tendency to move upwards, but apart from that, as already mentioned, the learned options are not very interpretable. Inspection reveals the policies do display a fair amount of diversity across options.

\begin{figure}[htb]
    \centering
    \includegraphics[width=\columnwidth]{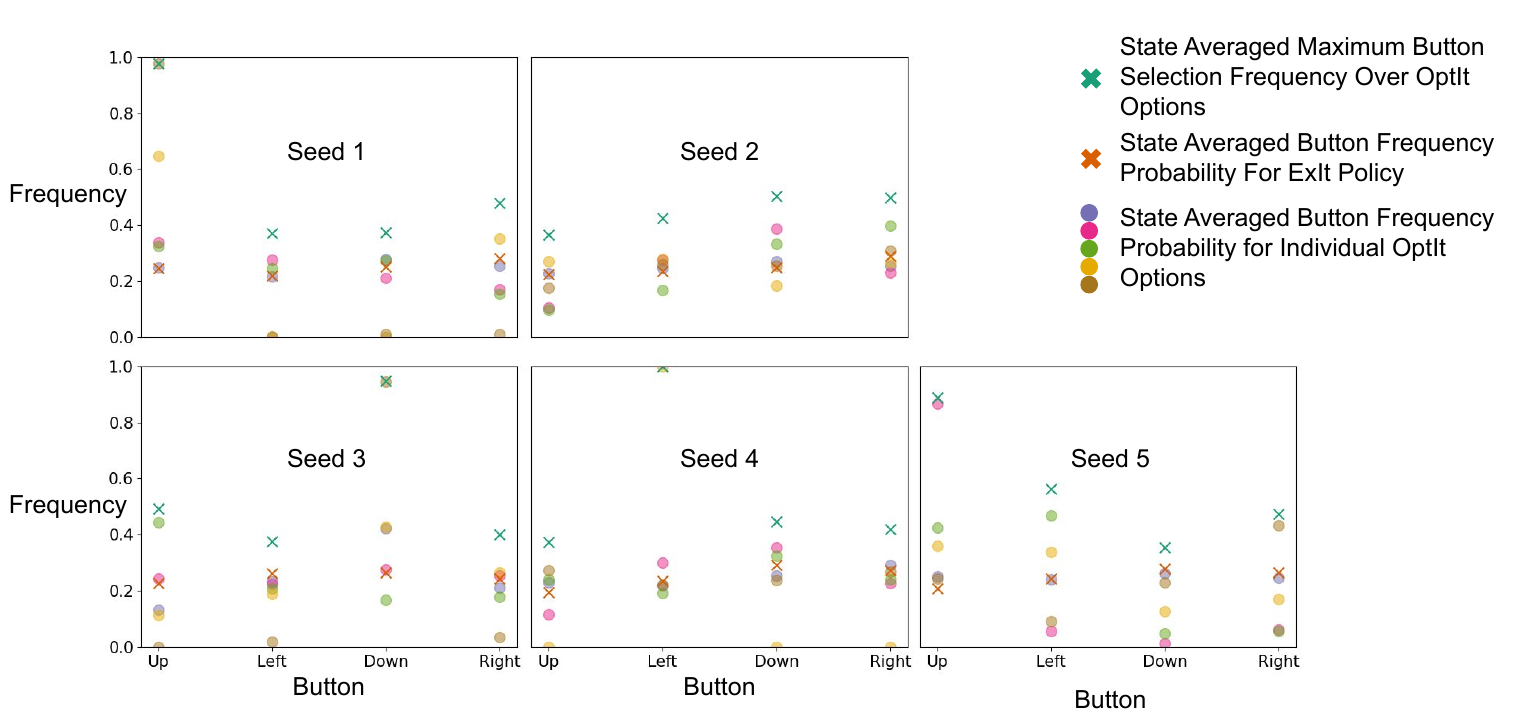}
    \caption{A visual comparison of the selection frequencies for each of the 4 buttons in the HierarchicalElectricProcMaze controller environment. The values were obtained by initializing the environment in a 1000 random positions, each in a random maze with the agent in the center of the controller environment.  The green X shows the maximum over options of the frequency with which each button was reached, averaged over states. The orange X similarly shows the frequency with which each button was reached by the single learned policy of ExIt. The colored dots show the state averaged button selection frequency for individual OptIt options. Each of 5 random seeds used for training are displayed separately.}
    \label{option_button_frequency}
\end{figure}

In order to gauge the degree of behavioural diversity among the learned options, we randomly initialize 1000 random environment states each in a random maze with the agent in the center of the controller environment. We then perform 1000 length 20 rollouts of each option in each of these random states and note the first button reached (Up, Down, Left or Right). We simply discard rollouts in which no button is reached within 20 steps. A visualization comparing the distribution of buttons reached by different options, as well as the single learned policy of ExIt is show in Figure~\ref{option_button_frequency}.

Note that in most cases (with the exception of Seed 2), we observe the emergence of at least one option that is sharply favors the selection of a particular button. However, unlike in the comparatively simple Compass environment in Section~\ref{sec:compass}, we do not observe an option specializing to each button across all states. On the other hand, the selection frequencies for the single learned ExIt policy are generally close to uniform as may be expected. Furthermore, looking at the state averaged maximum button selection frequency over options (green X) we see that when fixing the state there appears to be significantly more specialization among options than when we average over states. Note that, the green X is necessarily higher than all of the colored dots simply because the expectation of a maximum is always greater than or equal to the maximum of an expectation.

More quantitatively, we looked at an empirical estimate of the mutual information\footnote{To compute this we perform 1000 rollouts for each state-option combination for 1000 states, throwing out all rollouts which did not hit a button within 20 steps. Within the remaining rollouts we find the overall frequency $f_i$ with which each button was reached as well as $f_{i|n}$, the frequency of each button within those rollouts in which a particular option was used. We estimate the button entropy $\hat{H}(i)\approx \sum_i f_i\log(f_i)$ and option conditional button entropy $\hat{H}(i|n)\approx \sum_i f_{i|n}\log(f_{i|n})$. Finally we estimate the mutual information as $\hat{H}(i)-\frac{1}{N}\sum_n \hat{H}(i|n)$. We estimate the mean state conditional entropy and mutual information similarly except that the frequencies are computed separately for each state, and we compute a final average over the results across all 1000 sampled states.} between the option selected and the first button reached using the same setup outlined previously. The overall mutual information, without conditioning on state, was $0.19\pm 0.04$ compared to the total button selection entropy of $1.31\pm 0.02$. Hence, without conditioning on the state, the option selected is a poor predictor of the button reached. However, if we instead look at the mean state conditional mutual information we get $0.36\pm 0.04$ compared to the mean state conditional button entropy of $0.87\pm 0.03$. Hence, within a particular state, the selected option accounts for a little under half the entropy in the distribution of selected buttons.

Overall, while the learned options do not seem to facilitate straightforward interpretability in this case, they do present significant diversity and appear to be strongly associated with the button selected when conditioning on a particular starting state. This is likely a factor in the improved performance observed for 5 option OptIt compared to ExIt in HierarchicalElectricProcMaze.

\section{A Simple and Efficient Approach for Jointly Learning Termination Conditions, Option Policies and the Policy Over Options from Trajectory Data}\label{learned_option_termination}
In this work, we focus on the case where options are executed and learned for a fixed horizon. However, one could straightforwardly extend this method to the case where termination conditions are learned. Here, we outline an efficient algorithm for this. Our approach is functionally equivalent to the algorithm of~\citet{fox2017multi}. However, they propose an explicit forward-backward algorithm which is rather involved. Here we note that one can achieve the same thing by simply using the forward recursion relationship discussed by~\citet{fox2017multi} to compute the quantity we need to optimize and then computing the associated gradient using any automatic differentiation framework. We outline this approach explicitly since it may be useful to anyone trying to implement a similar idea.

For the purposes of this section, we assume we have a dataset of length $K$ trajectories $\tau=s_0,a_0,....,s_K$ which we assume are generated by using a search procedure to select actions at each step. We wish to maximize the log probability of the data under the distribution induced by a learned policy over options, option policies and a termination function. More precisely, we model the trajectories as generated by first using the policy over options $\rho(n|s_0;\theta)$ to choose an initial option, then selecting an action from $\pi_n(\cdot|s_k; \theta)$ for each step until the option terminates. At each step $k\geq1$, we also terminate the current option with probability given by the option termination function $\psi(s_k; \theta)$. If option termination does occur at step $k$, then we sample a new option from $\rho(n|s_k;\theta)$ and select actions for $j\geq k$ from $\pi_n(\cdot|s_j; \theta)$ until the next termination or the end of the trajectory. We seek to optimize the probability $\P(s_0,a_0,....,s_K;\theta)$ of trajectories in the dataset being generated by this process. Towards this goal, we begin with a recursion relation for $\phi_k(n;\theta)\defeq \P(s_0,a_0,....,s_k, n_k=n;\theta)$ outlined by~\citet{fox2017multi}, where $n_t$ denotes the option whose policy is used to select the action at time $t$. The recursion relation is as follows:
\begin{align*}
\phi_0(n;\theta)&=p_0(s_0)\rho(n|s_0;\theta)\\
    \phi_{k+1}(n;\theta)&=\begin{multlined}\left(\sum_{n^\prime} \phi_k(n^\prime;\theta)\pi_{n^\prime}(a_k|s_k;\theta)p(s_{k+1}|s_k,a_k;\theta)\psi_{n^\prime}(s_{k+1};\theta)\rho(n|s_{k+1};\theta)\right)\\+\phi_k(n;\theta)\pi_n(a_k|s_k;\theta)p(s_{k+1}|s_k,a_k)(1-\psi_{n}(s_{k+1};\theta)).
\end{multlined}
\end{align*}
Intuitively, the first term of the second equation accounts for the probability that some option terminated at time $k+1$ and option $n$ was selected from $\rho$ thereafter. The second term accounts for the probability that option $n$ was already being executed at time $k$ and did not terminate at $k+1$. Now note that we can factor out the transition probabilities, which are not under the agent's control, such that $\phi_k(n;\theta)=\tilde{\phi}_k(n;\theta)p_0(s_0)\prod_{t=0}^{k-1}p(s_{t+1}|s_t,a_t)$ with
\begin{align*}
\tilde{\phi}_0(n;\theta)&=\rho(n|s_0;\theta)\\
    \tilde{\phi}_{k+1}(n;\theta)&=\begin{multlined}\left(\sum_{n^\prime} \tilde{\phi}_k(n^\prime;\theta)\pi_{n^\prime}(a_k|s_k;\theta)\psi_{n^\prime}(s_{k+1};\theta)\right)\rho(n|s_{k+1};\theta)\\+\tilde{\phi}_k(n;\theta)\pi_n(a_k|s_k;\theta)(1-\psi_{n}(s_{k+1};\theta)).\numberthis\label{forward_recursion}
\end{multlined}
\end{align*}
We can then find the probability $\P(s_0,a_0,....,s_K;\theta)$ by simply marginalizing out the final option
\begin{align*}
\P(s_0,a_0,....,s_K;\theta)&=\sum_n\phi_K(n;\theta)\\
&= p_0(s_0)\prod_{t=0}^{K-1}p(s_{t+1}|s_t,a_t)\sum_n\tilde{\phi}_K(n;\theta).
\end{align*}
Taking the log of both sides we get 
\begin{equation*}
\log(\P(s_0,a_0,....,s_K;\theta))=\log\left(p_0(s_0)\prod_{t=0}^{K-1}p(s_{t+1}|s_t,a_t)\right)+\log\left(\sum_n\tilde{\phi}_K(n;\theta)\right)
\end{equation*}
Note that the left term is independent of $\theta$ hence we get
\begin{equation*}
\frac{\partial}{\partial\theta}\log(\P(s_0,a_0,....,s_K;\theta))=\frac{\partial}{\partial\theta}\log\left(\sum_n\tilde{\phi}_K(n;\theta)\right).
\end{equation*}
Having computed $\sum_n\tilde{\phi}_K(n;\theta)$ via the forward recursion in Equation~\ref{forward_recursion}, we can simply backpropagate to compute its gradient using any standard automatic differentiation framework.\footnote{In practice, one may want to implement the recursion in Equation~\ref{forward_recursion} in log-space, using a numerically stable implementation of LogSumExp to compute $\log(\tilde{\phi}_{k+1}(n;\theta))$ at each step.} The computation required for each step of the forward recursion is dominated by a sum over $N$ options, which can be computed once and shared across Equation~\ref{forward_recursion} for all $n$, and it's unrolled for $K$ steps hence the total complexity is on the order of $\mathcal{O}(KN)$, the order of the backward pass is the same. Note that the order of complexity is the same as computing the likelihood under a set of options and associated option policy even without termination conditions, so including termination conditions is no harder in that regard. However, without termination, we can parallelize over time steps which is no longer possible with termination conditions as we have to compute Equation~\ref{forward_recursion} sequentially.

The learned termination conditions, in combination with the learned option policies, could be utilized for planning in various ways. One simple approach would be to incorporate option termination into the rollouts. The MCS planner would select the initial option and action for each rollout, but at each subsequent step the option termination function would be queried and the current option terminated with probability $\psi_{n^\prime}(s_{k+1};\theta)$. When termination occurs in a rollout a new option would be selected according to $\rho$ as normal. This would allow for greater flexibility, as the learned options wouldn't necessarily need to be useful for the entirety of an arbitrary rollout horizon but could be switched mid-rollout based on the environment observations. For example, in HierarchicalElectricProcMaze it might be desirable to terminate the current option upon reaching one of the buttons in the controller-environment.

\section{Hyperparameters}\label{hyperparameters}
\begin{figure}[htb]
    \centering
    \includegraphics[width=\columnwidth]{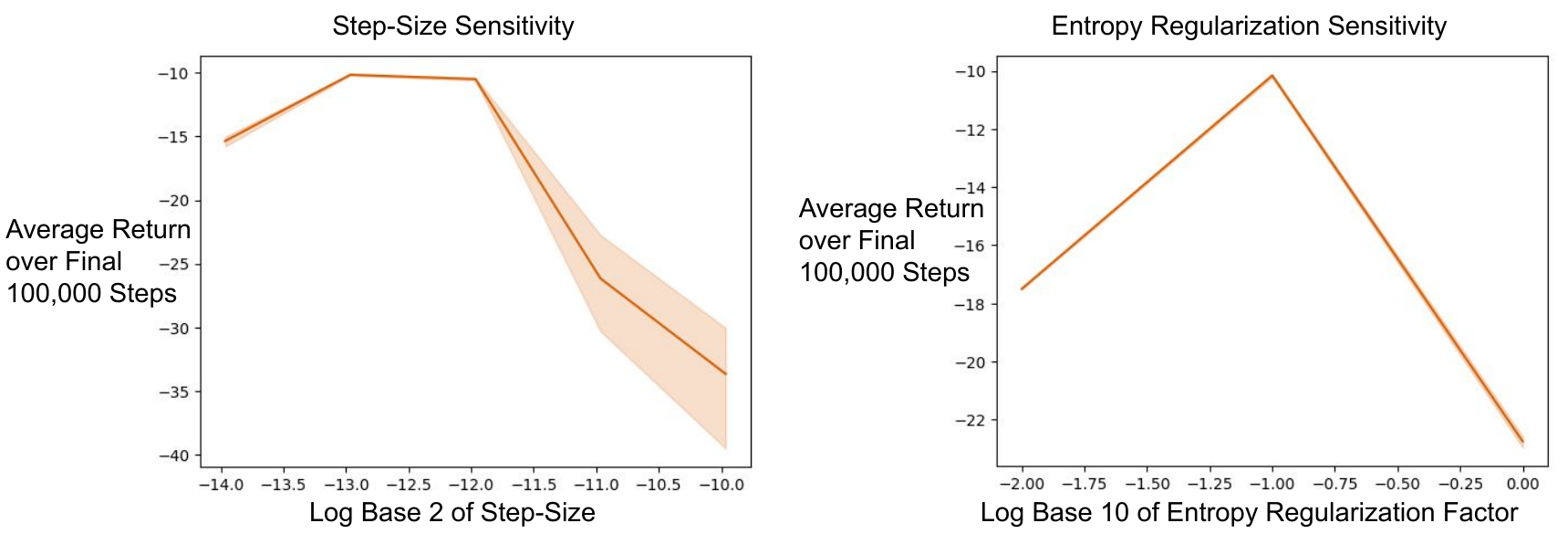}
    \caption{Sensitivity curves for ExIt resulting from grid sweep over step-size $\alpha$ and entropy regularization factor $\beta$ in ElectricProcMaze7. In each plot, the other hyperparameter is fixed to its best value from the grid-search while varying the hyperparameter of interest. Error bars show 95\% confidence interval over 5 random seeds.}
    \label{hyperparameter_sensitivity}
\end{figure}
Table~\ref{hyperparameter_table} shows the hyperparameters used in the Compass and ElectricProcMaze experiments. For ElectricProcMaze, the step-size parameter $\alpha$ and entropy regularization factor $\beta$ were tuned for the ExIt baseline from $\alpha\in\{0.0000625, 0.000125, 0.00025, 0.0005, 0.001\}$ and $\beta\in\{0.01, 0.1, 1.0\}$. We fix the option step-size (policy step-size for ExIt) to $\alpha$ and set the value step-size to $2\alpha$. We evaluate each hyperparameter combination over 5 random seeds and choose the one with the best average return over the last 100,000 of 500,000 time-steps. We display sensitivity curves resulting from this grid-search in Figure~\ref{hyperparameter_sensitivity}. Other hyperparameters were fixed to reasonable defaults.

For Compass, we mainly kept the same values from ElectricProcMaze with a few deliberate exceptions. We drastically reduced the simulations budget such there would be insufficient rollouts under a random policy to be likely to sample the optimal action sequence. We increased the option rollout length such that it was possible for rollouts to reach any edge of the grid from any starting location. We reduced the number of options to 4 to emphasize that 4 options were sufficient in this case. We also reduced $\beta$ to $0.01$ to facilitate convergence to near-optimal performance and cleaner learned options. At $\beta=0.1$, OptIt still performed significantly better than ExIt with a final average return of around $0.9$, and the learned options still showed a tendency to prefer one of the 4 cardinal directions each but became significantly more chaotic.

\begin{figure}[htb]
    \centering
    \includegraphics[width=0.5\columnwidth]{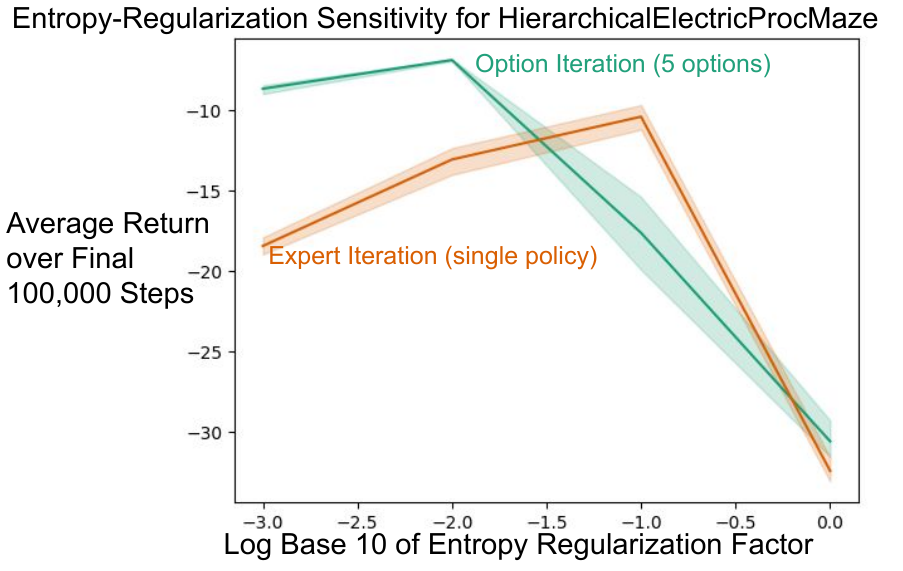}
    \caption{Sensitivity curve for ExIt and OptIt resulting from grid sweep entropy regularization factor $\beta$ in HierarchicalElectricProcMaze. OptIt achieves optimal performance at a lower entropy regularization and remains robust when lowering the regularization further. Error bars show 95\% confidence interval over 5 random seeds.}
    \label{ship_EPM5_beta_sensitivity}
\end{figure}
For HierarchicalElecticProcMaze, we also used most of the same hyperparameters from ElectricProcMaze. We increase the option rollout length to 8 to match the width of the control grid. We also doubled the number of hidden units as we found in preliminary experiments that this significantly improved the performance of ExIt indicating that the 400 hidden unit network was significantly underparameterized for the more challenging problem. Finally, we found it was nessesary to reduce the entropy regularization to observe a performance benefit for OptIt. To facilitate a fair comparison, and a more complete picture of the behaviour of each approach, we performed a sweep over $\beta\in\{0.001, 0.01, 0.1, 1.0\}$. The results of the sweep are displayed in Figure~\ref{ship_EPM5_beta_sensitivity}.
\begin{table}[H]
\centering
\begin{tabular}{|c|c|c|c|}
\cline{1-4}
\textbf{Hyperparameter} & \textbf{ElectricProcMaze}& \textbf{HierarchicalElectricProcMaze} & \textbf{Compass}\\
\cline{1-4}
Number of Hidden Layers&3&---&---\\
Number of Hidden Units&400&800&400\\
Hidden Activation& ELU&---&---\\
Optimizer&AdamW&---&---\\
Adam $\beta_1$&0.9&---&---\\
Adam $\beta_2$&0.99&---&---\\
Adam $\epsilon$&1e-5&---&---\\
Adam Weight Decay &1e-6&---&---\\
Option Step-Size& 1.25e-4 \textbf{(Tuned for ExIt)}&---&---\\
Value Function Step-Size& 2.5e-4 (Twice Above)&---&---\\
Running Average Variance Decay& 0.99&---&---\\
Discount Factor&0.99&---&---\\
Batch Size& 250&---&---\\
Number of Parallel Workers& 16&---&---\\
Gradient Updates per Environment Step& 16&---&---\\
Entropy Regularization Factor ($\beta$)& 0.1\textbf{(Tuned for ExIt)}& 0.1 (ExIt)/0.01 (OptIt)&0.01\\
Simulation Budget & 1000& ---& 50\\
Option Rollout Length ($K$)& 5& 8& 20\\
Number of Options & 5& ---& 4\\
Buffer Size&100,000&---&---\\
Training Start Time&100&---&---\\
\cline{1-4}
\end{tabular}
\caption{Table of hyperparameters used in Compass and ElectricProcMaze experiments. Dashes denote that the same value is used as ElectricProcMaze.}\label{hyperparameter_table}
\end{table}

\end{document}